%% file: main.tex
\documentclass[12pt,final]{l4dc2022}
\makeatletter
\def\set@curr@file#1{\def\@curr@file{#1}} 
\makeatother
\input{preamble}
\usepackage[normalem]{ulem}
\usepackage{cancel}

\title[Safety-Aware Preference-Based Learning for Safety-Critical Control]{
Safety-Aware Preference-Based Learning for Safety-Critical Control 
}
\usepackage{times}

\newcommand{\K}{\mathcal{K}}

\usepackage{bm} 
\DeclareMathOperator*{\argmax}{argmax}
\renewcommand{\argmax}{\operatornamewithlimits{argmax}} 

\author{%
 \Name{Ryan K. Cosner}$^1$      \Email{rkcosner@caltech.edu} \\ 
 \Name{Maegan Tucker}$^1$       \Email{mtucker@caltech.edu}\\
 \Name{Andrew J. Taylor}$^1$    \Email{ajtaylor@caltech.edu}\\
 \Name{Kejun Li}$^1$            \Email{kli5@caltech.edu}\\
 \Name{Tamas G. Molnar}$^1$     \Email{tmolnar@caltech.edu}\\
 \Name{Wyatt Ubellacker}$^1$    \Email{wubellac@caltech.edu}\\
 \Name{Anil Alan}$^2$           \Email{anilalan@umich.edu}\\
 \Name{Gábor Orosz}$^2$         \Email{orosz@umich.edu}\\
 \Name{Yisong Yue}$^{1,3}$          \Email{yyue@caltech.edu}\\
 \Name{Aaron D. Ames}$^1$       \Email{ames@caltech.edu}\\
 \addr $^1$ California Institute of Technology, Pasadena, CA, USA \\
 \addr $^2$ University of Michigan, Ann Arbor, MI, USA\\
 \addr $^3 $ Argo AI, Pittsburgh PA, USA
 \vspace{-15pt}
}

\begin{document}
\maketitle

\vspace{-15pt}
\begin{abstract} 
Bringing dynamic robots into the wild requires a tenuous balance between performance and safety. Yet controllers designed to provide robust safety guarantees often result in conservative behavior, and tuning these controllers to find the ideal trade-off between performance and safety typically requires domain expertise or a carefully constructed reward function. 
This work presents a design paradigm for systematically achieving behaviors that balance performance and robust safety by integrating \textit{safety-aware} Preference-Based Learning (PBL) with Control Barrier Functions (CBFs).
Fusing these concepts---safety-aware learning and safety-critical control---gives a robust means to achieve safe behaviors on complex robotic systems in practice. We demonstrate the capability of this design paradigm to achieve safe and performant perception-based autonomous operation of a quadrupedal robot both in simulation and experimentally on hardware.
\end{abstract}


\vspace{-10pt}

\section{Introduction}

\input{Sections/Intro}

\section{Safety-Aware Preference-Based Learning}
\input{Sections/PBL}

\section{Robust Safety-Critical Control}
\input{Sections/AT_SecIII_Draft_v2}

\section{Integrating Safety-Aware Preference-Based Learning with Safety-Critical Control}
\input{Sections/AT_SecIV_Draft}

\section{Experimental Results}
\input{Sections/experiments}


\clearpage

\acks{
We thank the anonymous reviewers for helpful feedback.
This research is generously supported in part by the National Science Foundation (CPS Award \#1932091 and GRFP Award DGE‐1745301), Dow (\#227027AT), Wandercraft, BP p.l.c., AeroVironment, and the ZEITLIN Funds. 
}

\bibliography{taylor_main, other}
 \newpage

\appendix

\section{Proof of Theorem \ref{thm:all_uncertainties}} \label{apdx:thm_proof}
\input{Sections/Appendices/theoremProof}


\end{document}

%% file: preamble.tex
\usepackage{hyperref}
\usepackage{graphicx}		
\usepackage{wrapfig}
\usepackage[format=plain,font=footnotesize,labelfont=bf,labelsep=period]{caption}
\usepackage[export]{adjustbox}
\usepackage{bm}

\usepackage{amsmath}
\usepackage{amssymb}
\usepackage{mathtools}
\usepackage{algorithm}
\usepackage{algpseudocode}

\usepackage{pdfsync}
\usepackage[normalem]{ulem}
\usepackage{paralist}	
\usepackage[space]{grffile} 

\usepackage{color}

\newtheorem*{theorem*}{Theorem}
\newtheorem*{lemma*}{Lemma}
\newenvironment{proof}{\textit{Proof.}}{\hfill$\square$}

\newcommand{\R}{\mathbb{R}}

\newcommand{\C}{\mathcal{C}}

\definecolor{darkblue}{RGB}{0,0,102}
\definecolor{lightblue}{RGB}{77,77,148}

\definecolor{gold}{RGB}{234, 170, 0}
\definecolor{metallic_gold}{RGB}{139, 111, 78}

\newcommand{\mb}[1]{\mathbf{ #1 }}
\newcommand{\bs}[1]{\boldsymbol{ #1 }}

\newcommand{\derp}[2]{\frac{\partial #1 }{\partial #2 }}

\newcommand{\newpbl}{\textcolor{black}{SA-LineCoSpar}}
\newcommand{\newpblname}{\textcolor{black}{Safety-Aware LineCoSpar}}
\newcommand{\newpblfull}{\textcolor{black}{\newpblname~(\newpbl)}}

\newcommand{\lmat }{\begin{bmatrix}}
\newcommand{\rmat}{\end{bmatrix}}

\newcommand{\N}{\mathcal{N}}
\newcommand{\ord}{O}
\newcommand{\data}{\mathbf{D}}
\newcommand{\util}{r}
\newcommand{\utilvec}{\mb{\util}}
\newcommand{\vis}{V}
\renewcommand{\line}{L}
\newcommand{\act}{\mb{a}} 
\newcommand{\actspace}{A} 
\renewcommand{\data}{D} 
\renewcommand{\P}{\mathcal{P}} 

\newcommand{\issfparam}{\varphi}
\newcommand{\contbound}{\mu}

%% file: Sections/Intro.tex
The increasing demands of modern engineering problems have required a commensurate increase in the complexity of the underlying control systems being used. The process of designing these complex control systems is often accomplished by separating the design into individual subsystems such as sensing, planning, and low-level control, which are later integrated. A principal challenge in the integration of such complex systems is balancing safety with performance at the system level. When each individual subsystem is designed using over-approximations of worst-case scenarios, the integrated system becomes extremely conservative and exhibits poor performance \citep{singletary2021comparative, alan2021safe}. The commonly employed alternative is tuning the safety-performance trade-off of each component to achieve the desired system-level behavior \citep{ma2017bipedal}, which can be challenging even for domain experts as the tuning is often done via qualitative assessments.



For instance, for complex safety-critical systems, Control Barrier Functions (CBFs) have become a popular tool for the constructive synthesis of model-based controllers that endow nonlinear systems with rigorous guarantees of safety \citep{ames2014control, ames2019control, hobbs2021run}. As these safety guarantees are susceptible to inaccuracies in the models of a system's dynamics, actuators, and sensors, approaches have been proposed to deal with model uncertainty \citep{wang2018safe, taylor2019adaptive, castaneda2020gaussian, taylor2020learning}, disturbances \citep{jankovic2018robust, kolathaya2018input, clark2019control, santoyo2019verification, alan2021safe, choi2021robust}, and measurement errors \citep{takano2018robust, dean2020guaranteeing, cosner2021measurement}. These approaches can work well when deployed independently, but can be extremely conservative systems when used in conjunction. In practice, achieving performant behaviors with these methods is accomplished by conceding theoretical safety guarantees 
and tuning controller robustness parameters.





\begin{figure}[t]
    \centering
    \includegraphics[width=1\linewidth]{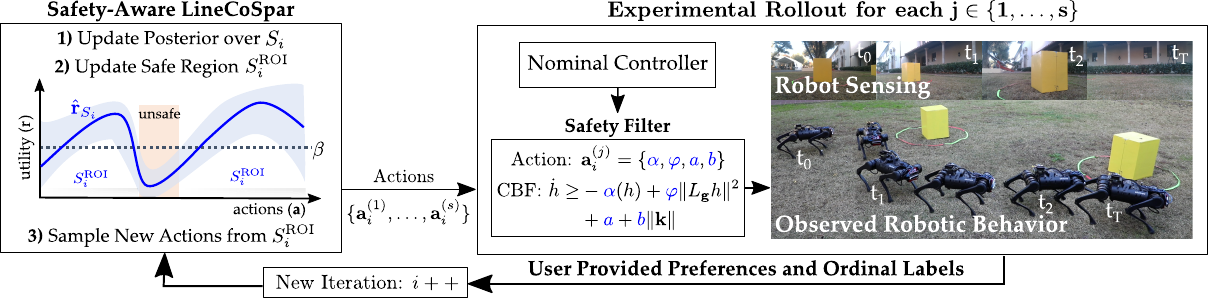}
    \caption{ An overview of the Safety-Aware Preference-Based Learning design paradigm. Safety-Aware LineCoSpar is used to generate actions which are rolled out in experiments as parameters of the CBF-based safety filter to obtain user preferences and safety ordinal labels which are then used to update the user's estimated utility and generate new actions.}
    \label{fig:overview}
    \vspace{-0.85cm}
\end{figure}

To reduce the burden on experts in controller tuning, we seek to incorporate Preference-Based Learning (PBL) into the design of safety-critical control systems. PBL has shown to be a powerful tool for converting subjective user preferences of system behavior (e.g., behavior A
is preferred over 
behavior B) into quantitative adjustments to design parameters. The main advantage of online PBL is its ability to interactively infer a user's latent utility function using only subjective feedback such as pairwise preferences and ordinal labels  \citep{yue2012k,shivaswamy2012online}. 
This methodology has been demonstrated in application for exoskeleton gait optimization \citep{tucker2020preference}, bipedal locomotion \citep{tucker2021preference}, spinal cord stimulation \citep{sui2018stagewise},  trajectory planning \citep{sadigh2017active,biyik2020active,jain2015learning}, search engines \citep{raman2013stable}, and recommender systems \citep{de2009preference}. For applications with actions that may be classified as safe or unsafe, \textit{safety-critical} PBL algorithms have been demonstrated to prevent unsafe actions from being sampled \citep{sui2015safe,sui2018stagewise,berkenkamp2016safe}. However, these safety-critical algorithms require worst-case approximations 
which may cause performant and safe actions to be characterized as catastrophically unsafe. Thus, we seek to formulate a \textit{safety-aware} approach to PBL that generally avoids unsafe actions without being overly conservative. 

In this work we propose a design paradigm for synthesizing performant and robust safety-critical controllers on real systems via safety-aware online PBL (illustrated in Fig.~\ref{fig:overview}). The contributions of this work are threefold. First, we propose \newpblfull, a modified version of LineCoSpar \citep{tucker2020human} capable of high-dimensional preference-based Bayesian optimization while also accounting for safety.
Second, we combine the robustness properties of Measurement-Robust CBFs (MR-CBFs) \citep{dean2020guaranteeing} to measurement uncertainty and Input-to-State Safe CBFs (ISSf-CBFs) \citep{kolathaya2018input} to disturbances with
reduced-order multi-layer safety-critical control \citep{molnar2021model} to
achieve provable safety guarantees in a parametric form amenable to \newpbl. Finally, we deploy these two methods together as a design paradigm for a safety-critical controller on a quadrupedal robot in simulation and on hardware in laboratory and outdoor settings. Additionally, this work is the first time that PBL has been used to tune a CBF-based controller, and the first time these CBF methods have been combined.




%% file: Sections/PBL.tex
\vspace{-0.5mm}

Preference-Based Learning (PBL) provides an approach for searching complex parameter spaces via subjective feedback, without an explicitly defined reward function. This is particularly relevant for safety-critical systems, as quantifying the user-preferred trade-off between robustness and performance is difficult. Moreover, poorly defined reward functions often result in “reward hacking” \citep{amodei2016concrete}, in which undesirable actions achieve high rewards. Here, we propose \newpblname\,(\newpbl), outlined in Alg.\,\ref{alg:episodic_learning}. This is a modification of the LineCoSpar algorithm \citep{tucker2020human},  which iteratively selects actions to query user for subjective feedback and updates its belief of the user's underlying utility function via Bayesian inference. 


\underline{\textit{Problem Setup:}}
Let $\act$ denote an action, such as a collection of $l$ parameters used in a feedback controller, that takes values in a finite search space $\actspace\subset\R^l$. We assume that each action $\act \in \actspace$ has an unknown utility to the user, defined by a function $\util :\actspace \to\R$. These utilities are given by
$\utilvec_{\actspace} = [\util(\act_1), \dots, \util(\act_{|\actspace|})]^ \top\in\R^{|\actspace|}$. In each iteration, $s\in\mathbb{N}$ actions are sampled from $\actspace$ and executed. Then, the user is queried for two forms of feedback: pairwise preferences and ordinal labels, describing \textit{performance} and \textit{safety}, respectively. This feedback is collected into dataset $\data$.

\underline{\textit{Modeling the Utility Function:}}
Since collecting an exhaustive dataset to estimate the unknown utility $\utilvec_{A}$ is expensive for non-trivial action spaces, we use Bayesian optimization (BO), a sampling efficient paradigm for identifying the optimizer. In BO, $\utilvec_{A}$ is modeled as a Gaussian process with prior $\N(\bm{0},\bs{\Sigma}^{\text{pr}})$, where each element of the covariance matrix $\bs{\Sigma}^{\text{pr}}\in\mathbb{S}^{|\actspace|\times|\actspace|}_{\succ 0}$ is computed as $\bs{\Sigma}^{\text{pr}}_{ij} = {k}(\act_i,\act_j)$ with a kernel function $k: \actspace \times \actspace \to \R$ and $\act_i\in A$ denoting the $i^\textup{th}$ action in $\actspace$. We select $k$ to be the squared exponential kernel, yielding a prior given by the multivariate Gaussian:
\begin{align}
    \P(\utilvec_{\actspace}) = \frac{1}{(2\pi)^{\lvert \actspace \rvert /2} \lvert \bs{\Sigma}^{\text{pr}}\rvert^{1/2}} \text{exp} \left(-\frac{1}{2}\utilvec_{\actspace}^\top{\left(\bs{\Sigma}^{\text{pr}}\right)}^{-1}\utilvec_{A}\right).
\end{align}
Given a dataset $\data$, the posterior is proportional to the likelihood and the prior by Bayes' theorem, i.e., $\P(\utilvec_{\actspace} \mid \data) \propto \P(\data \mid \utilvec_{\actspace})\P(\utilvec_{A})$.
We denote the maximum a posteriori (MAP) estimate of the posterior by $\hat{\utilvec}_{\actspace}\in\R^{|\actspace|}$, which is defined as $\hat{\utilvec}_{\actspace} \triangleq \argmax_{\utilvec_{\actspace} \in \R^{|\actspace|}} \P(\utilvec_{\actspace} \mid \data)$, noting that $\hat{\utilvec}_{\actspace}$ is equivalent to the minimizer of $\mathcal{S}(\utilvec_{\actspace}) = -\ln(\P(\data \mid \utilvec_{\actspace})) + \frac{1}{2} \utilvec_{\actspace}^T \left(\Sigma^{\text{pr}}\right)^{-1}\utilvec_{\actspace}$. As is common in BO, we model the posterior as a multivariate Gaussian centered at $\hat{\utilvec}_{\actspace}$ with the covariance $\bs{\Sigma}_{\actspace}\in \mathbb{S}^{|\actspace|\times|\actspace|}_{\succ 0} $ defined as $\bs{\Sigma}_{\actspace} = (\frac{\partial^2 \mathcal{S}}{\partial \utilvec_{\actspace}^2}(\hat{\utilvec}_{\actspace}))^{-1}$ \citep{chu2005preference}\footnote{
This is known as the Laplace approximation of the distribution $\P(\utilvec_{\actspace} \mid \data)$, i.e., $\P(\utilvec_{\actspace} \mid \data) \approx \N(\hat{\utilvec}_{\actspace}, \bs{\Sigma}_{\actspace})$.}. Additionally, we can improve tractability of calculating $\hat{\utilvec}_{\actspace}$ by reducing the action space $A$ to a subset $S \subset  A $, 
forming a partial characterization of the utilities denoted by $\P(\utilvec_S \mid \data) \approx \N(\hat{\utilvec}_S, \bs{\Sigma}_S)$, with $\utilvec_S,\hat{\utilvec}_S  \in \R^{|S|}$.

\underline{\textit{Preference Likelihood Function:}}
A pairwise preference is defined as a relation between two actions $\act_1,\act_2\in\actspace$, where $\act_1\succ\act_2$ if action $\act_1$ is preferred to $\act_2$. Since user preferences are expected to be corrupted by noise, we model individual pairwise preferences via a likelihood function: 
\begin{equation}
    \P(\act_1 \succ \act_2 \vert 
    \util(\act_1), \util(\act_2)) = g_p\left( \frac{\util(\act_1) - \util(\act_2)}{c_p}\right),
\end{equation}
where $g_p:\R \to [0,1]$ is any monotonically-increasing link function, and $c_p\in\R_{>0}$ accounts for preference noise. We select $g_p$ to be the sigmoid function, i.e., $g_p(x) = 1/(1 + e^{-x})$. 
Assuming conditional independence, the likelihood function for a collection of $K\in \mathbb{N}$ preferences, $\data_p$, can be modeled as the product of each individual preference likelihood: 
\begin{equation}
    \P(\data_p\vert \util(\act_{11}),\util(\act_{12}),\cdots,\util(\act_{K2})) = \prod_{k=1}^K \P(\act_{k1} \succ \act_{k2} \vert \util(\act_{k1}), \util(\act_{k2})),
\end{equation}
where $\act_{k1},\act_{k2}\in \actspace$ are the preferred and non-preferred actions, respectively, in the $k^{th}$ preference. 

\underline{\textit{Ordinal Likelihood Function:}}
 We partition the action space into ``unsafe'' and ``safe'' actions by leveraging the ordinal nature of these definitions (i.e., unsafe actions are always considered worse than safe actions). A user provides this feedback as an ordinal label, which assigns an action to a discrete ordered category such as ``bad" and ``good" \citep{chu2005gaussian}. While ordinal labels can be generalized to any number of ordinal categories (c.f. \cite{li2021roial}), we utilize just two categories to represent ``unsafe'' and ``safe''. In this case, the action space is decomposed into two disjoint sets, $\actspace = \ord_1\cup\ord_2$, with $\act\in\ord_1$ if $\util(\act)<\beta$ and $\act\in\ord_2$ if $\util(\act)\geq \beta$, with the ordinal threshold $\beta\in\R$. As with preferences, we assume that ordinal label feedback is corrupted by noise and is modeled as:
\begin{align}
    \P(\mb{a}\in \ord_1 \mid \util(\act)) = g_o\left( \frac{\beta-\util(\act)}{c_{o}}\right), \qquad  \P(\mb{a}\in \ord_2 \mid \util(\act)) = 1-g_o\left( \frac{\beta-\util(\act)}{c_{o}}\right),
\end{align}
where $g_o: \R \to [0,1]$ is any  monotonically-increasing link function and $c_o$ quantifies the noise in the ordinal label feedback. Again, we select $g_o$ to be the sigmoid function $g_o(x) = 1/(1 + e^{-x})$. Assuming conditional independence of ordinal label queries, the likelihood function for a collection of $M\in\mathbb{N}$ ordinal labels, $\data_o$, can be modeled as the product of each individual ordinal likelihood:
\begin{align}
    \P(\data_{o} \mid \util(\act_1),\cdots,\util(\act_k)) = & \prod_{k=1}^{M} \P\left(\mb{a}_k\in\ord_{o(k)} \mid \util(\act_k)\right),
\end{align}
where $\act_k\in A$ refers to the action corresponding to the $k^\textup{th}$ ordinal label, $o(k)\in\{1,2\}$. For our simulation and experiments, the hyperparameters $c_p$, $c_o$, $\beta$ are determined in advance. Lastly, assuming conditional independence of the feedback mechanisms, the combined likelihood function is calculated as the product of the individual likelihoods, $\P(\data \mid \util) = \P(\data_p \mid \util)\P(\data_o \mid \util)$.


\underline{\textit{Sampling New Actions:}}
In the first iteration ($i=1$), $s\in\mathbb{N}$ actions are sampled randomly from $\actspace$, recorded as the set of visited actions $\vis_1 = \{ \act_1^{(1)},\dots,\act_1^{(s)}\}$, executed on the system, and the preferences and ordinal labels are collected into a dataset $D_1$. In each subsequent iteration ($i > 1$), $s$ new actions are sampled using Thompson sampling, which is shown to have desirable regret minimization properties \citep{chapelle2011empirical}. Ideally, Thompson sampling draws $s$ samples from the posterior $\P(\utilvec_{\actspace} \mid \data_{i-1})$, i.e $\utilvec^{(j)} \sim \mathcal{P}(\utilvec_{\actspace} \mid \data_{i-1})$ for $j \in \{ 1, \dots, s \}$, and the action  $\act_i^{(j)}\in \actspace$ maximizing each $\utilvec^{(j)}$ is selected to execute on the system. These sampled actions $\{\act_i^{(1)},\dots,\act_i^{(s)}\}$ are concatenated with $\vis_{i-1}$ to produce $\vis_i$, executed on the system, and the resulting preferences and ordinal labels are concatenated with $\data_{i-1}$ to produce $\data_{i}$. However, since it is intractable to approximate $\mathcal{P}(\utilvec_{\actspace} \mid \data)$ for high-dimensional action spaces, we utilize a dimensionality-reduction technique introduced in \cite{tucker2020human} that instead updates the posterior over a subset $S_i \subset \actspace$. Motivated by \cite{kirschner2019adaptive}, we construct the subset as $S_i = \line_i \cup \vis_{i-1}$, where $\line_i \subset \actspace$ is the collection of $e\in\mathbb{N}$ actions in $\actspace$ closest to a randomly drawn line $\ell_i\subset\R^l$. This line is drawn to intersect with the believed best action, computed as $\hat{\act}_{i-1}^* = \argmax_{\act \in \vis_{i-1}} \hat{\utilvec}_{\vis_{i-1}}(\act)$ where $\hat{\utilvec}_{\vis_{i-1}}$ is the MAP estimate of the posterior $\mathcal{P}(\utilvec_{\vis_{i-1}} \mid \data_i)$. See \cite{tucker2020human} for more details.

\underline{\textit{Safety-Aware Sampling:}}
It is important to avoid unsafe actions during sequential decision making in certain applications, such as learning robotic controllers on hardware, where low-reward actions might lead to physical damage of the platform. Safe exploration algorithms \citep{sui2015safe, sui2018stagewise, berkenkamp2016safe} considered the setting where actions below a prespecified safety threshold are catastrophic and must be avoided at all cost. In our work, since we construct controllers that account for safety, we adopt a more optimistic learning approach called \textit{safety-aware}. In this case, actions labeled by a human as ``unsafe'' are not catastrophic but undesirable. Thus, the algorithm \textit{avoids} these actions; whereas the safe exploration algorithms guarantee that no such actions are sampled which can be sometimes exceedingly conservative in settings like ours.

To achieve this safety-awareness, we leverage the approach introduced in \cite{li2021roial}, which uses ordinal labels to identify a \textit{region of interest} (ROI) in $\actspace$. In this work, the ROI is defined to be the actions labeled as ``safe''. In each iteration $i$ we estimate an ROI within the set $S_i$ as:
\begin{align}
    S^{\text{ROI}}_i = \{\act \in S_i \mid \hat{\utilvec}_{S_i}(\act) + \lambda \bs{\sigma}_{S_i}(\act) > \beta\},
\end{align}
where $\hat{\utilvec}_{S_i}(\act)$ and $\bs{\sigma}_{S_i}(\act)$ are the posterior mean and standard deviation, respectively, evaluated at the action $\act \in S_i$. The variable $\lambda \in \R$ determines how conservative the algorithm would be in estimating the safety region, as illustrated in Figure \ref{fig:sa-linecospar}. We see that lower values of $\lambda$ result in fewer unsafe actions being sampled, with only a slight effect on sample-efficiency.
The restriction to $S_i^{\text{ROI}}$ is added to LineCoSpar by only considering actions in $S_i^{\text{ROI}}$ during Thompson sampling. We refer to this as \newpblname\, (\newpbl), with the full algorithm outlined in Alg. \ref{alg:episodic_learning}. 

\noindent
\adjustbox{valign=t}{\begin{minipage}[t]{0.5\textwidth}
\begin{algorithm2e}[H]
\DontPrintSemicolon
\caption{\newpblname}
\label{alg:episodic_learning}
 \SetKwInOut{Input}{input}
 \Input{$s$ uniform random actions ($\vis_1 \subset \actspace$), corresponding feedback ($\data_1$), }
 \begin{small} 
    \For{$i=2,\dots,N$}{
    Update posterior over $\vis_{i-1}$\;
    $\hat{\act}^*_{i-1} \leftarrow \argmax_{\act \in \vis_{i-1}} \hat{\utilvec}_{\vis_{i-1}}(\act)$ \;
    $\line_i \leftarrow$ New linear subspace intersecting $\hat{\act}^*_{i-1}$\;
    Construct subspace $S_i = \line_i \cup \vis_{i-1}$ \;
    Update the model posterior over $S_i$ \;
    Determine region of interest $S^{\text{ROI}}_i$ \;
    \For{$j = 1,\dots,s$}{
        $\util^{(j)} \sim \N(\hat{\utilvec}_{S_i},\bs{\Sigma}_{S_i})$ \;
        $\act_i^{(j)} \leftarrow \argmax_{\act \in S^{\text{ROI}}_i}\util^{(j)}$\;
    }
    Deploy $\{\act_i^{(1)},\dots \act_i^{(s)}\}$ on system \;
    $\vis_i \leftarrow \vis_{i-1} \cup \{\act_i^{(1)},\dots \act_i^{(s)}\}$ \;
    $\data_i \leftarrow \data_{i-1} \cup \text{new prefs.} \cup \text{new ord. labels}$ \;
} 
\end{small}
\end{algorithm2e}
\end{minipage}}
\hfill
\adjustbox{valign=t}{
\begin{minipage}[t]{0.45\textwidth}
    \vspace{0mm}
    \includegraphics[width=\linewidth]{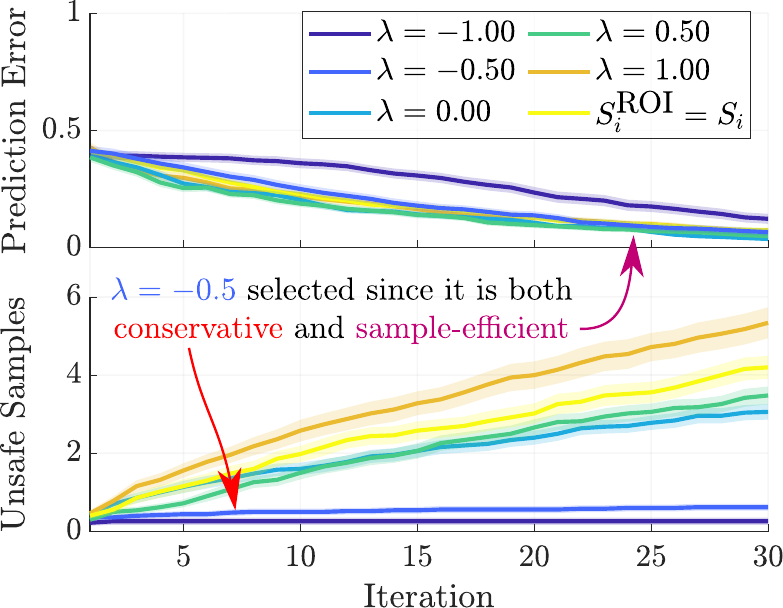}
    \captionof{figure}{A comparison of \newpbl~ and standard LineCoSpar on a synthetic utility function (drawn from the Gaussian prior) averaged over 50 runs with standard error shown by the shaded region. The safety-aware criteria reduces the number of sampled unsafe actions with a minimal effect on the prediction error, defined as $\lvert \hat{\act}_i^* - \act^* \rvert$ with $\hat{\act}_i^* \triangleq \argmax_{\act} \hat{\utilvec}_{S_i}$ and $\act^* \triangleq \argmax_{\act} \util(\act)$.}
    \label{fig:sa-linecospar}
\end{minipage}}

%% file: Sections/AT_SecIII_Draft_v2.tex
In this section, we formalize robust safety and discuss safe controller synthesis through the use of Control Barrier Functions (CBFs), that ultimately yield controllers whose parameters are to be updated with \newpbl.
Consider the following nonlinear control-affine system:
\begin{equation}
    \dot{\mb{x}} = \mb{f}(\mb{x}) + \mb{g}(\mb{x})(\mb{v}+\mb{d}(t)),
    \label{eq:reduced_model}
\end{equation}
with state ${\mb{x} \in \R^{n}}$,
input ${\mb{v} \in \R^{m}}$, functions ${\mb{f}:\R^{n}\to\R^{n}}$ and ${\mb{g} : \R^{n}\to\R^{n \times m}}$ assumed to be locally Lipschitz continuous on their domains, and piecewise continuous disturbance signal $\mb{d}:\R_{\geq 0} \to \R^{m}$ for which we define $\Vert\mb{d}\Vert_{\infty} \triangleq \sup_{t \geq 0} \Vert\mb{d}(t)\Vert$. Specifying the input via a controller $\mb{k}:\R^{n}\to \R^{m}$ that is locally Lipschitz continuous on its domain yields the closed-loop system:
\begin{equation}
    \dot{\mb{x}} = \mb{f}(\mb{x}) + \mb{g}(\mb{x})(\mb{k}(\mb{x})+\mb{d}(t)).
    \label{eq:closed_loop_system}
\end{equation}
We assume for any initial condition ${\mb{x}(0) = \mb{x}_0 \in \R^{n}}$ and disturbance $\mb{d}$, this system has a unique solution $\mb{x}(t)$ for all $t \in \R_{\geq 0}$. We consider this system safe if its state $\mb{x}(t)$ remains in a \textit{safe set} $\mathcal{C} \subset \R^{n}$, defined as the 0-superlevel set of a continuously differentiable function $h:\R^{n}\times\R^p\to\R$: 
\begin{equation}
    \mathcal{C} = \{ \mb{x} \in \R^{n} : h(\mb{x},\bs{\rho}) \geq 0 \}, \label{eq:safe_set}
\end{equation}
where $\bs{\rho}\in\R^p$ are constant application-specific parameters. We say the set $\mathcal{C} \subset \R^{n}$ is \textit{forward invariant} if for every $\mb{x}_0 \in \mathcal{C}$ the solution $\mb{x}(t)$ to \eqref{eq:closed_loop_system} satisfies $\mb{x}(t) \in \mathcal{C}$ for all $t \geq 0 $. The system \eqref{eq:closed_loop_system} is \textit{safe} with respect to $\mathcal{C}$ if $\mathcal{C}$ is forward invariant. Ensuring the safety of the set $\C$ in the absence of disturbances and measurement error can be achieved through {\em Control Barrier Functions (CBFs)}:
\begin{definition}[Control Barrier Functions (CBF) \citep{ames2014control}]
The function $h$ is a Control Barrier Function (CBF) for \eqref{eq:reduced_model} on $\C$ if there exists $\alpha\in\K_{\infty}^{\rm e}$\footnote{We say that a continuous function ${\alpha : \R_{\geq 0} \to \R_{\geq 0}}$ is {\em class~$\mathcal{K}_\infty$} (${\alpha \in \mathcal{K}_\infty}$) if ${\alpha(0) = 0}$, $\alpha$ is strictly monotonically increasing, and $\lim_{r\to\infty}\alpha(r) = \infty$. We say that a continuous function $\alpha : \R \to \R$ is {\em class~$\K_{\infty}^{\rm e}$} (${\alpha \in \K_{\infty}^{\rm e}}$) if ${\alpha(0) = 0}$, $\alpha$ is strictly monotonically increasing, $\lim_{r\to\infty}\alpha(r) = \infty$, and $\lim_{r\to-\infty}\alpha(r) = -\infty$.}  such that for all $\mb{x}\in\R^n$:
\begin{equation}
    \label{eq:cbf_constraint}
     \sup_{\mb{v} \in \R^{m}}
        \underbrace{\derp{h}{\mb{x}}(\mb{x},\bs{\rho})\mb{f}(\mb{x})}_{L_\mb{f}h(\mb{x},\bs{\rho})} + \underbrace{\derp{h}{\mb{x}}{(\mb{x},\bs{\rho})}\mb{g}(\mb{x})}_{L_\mb{g}h(\mb{x},\bs{\rho})}\mb{v} > -\alpha(h(\mb{x},\bs{\rho})).
\end{equation}
\end{definition}

While it may be possible to synthesize controllers that render a given set $\C$ safe in the presence of disturbances \citep{jankovic2018robust}, this may result in overly-conservative behavior. Instead, we consider how safety properties degrade with disturbances via the following definition.
\begin{definition}[Input-to-State Safety \citep{kolathaya2018input}]
\label{def:issf}
   The system \eqref{eq:closed_loop_system} is Input-to-State Safe (ISSf) with respect to $\mathcal{C}$ if there exists $\gamma \in \mathcal{K}_\infty$ such that for all $\delta\in\R_{\geq 0}$ and disturbances $\mb{d}:\R_{\geq 0}\to\R^m$ satisfying $\Vert\mb{d}\Vert_\infty\leq\delta$, the set $\mathcal{C}_{\delta}\subset\R^n$ defined as: 
    \begin{align}
       \mathcal{C}_{\delta} = \{\mb{x} \in \R^{n} : h(\mb{x}, \bs{\rho}) \geq - \gamma(\delta)\},
    \end{align}
    is forward invariant. The function $h$ is an Input-to-State Safe Control Barrier Function (ISSf-CBF) for~(\ref{eq:reduced_model}) on $\C$ with parameter $\issfparam\in\R_{\geq 0}$ if there exists $\alpha\in\K_{\infty}^{\rm e}$ such that for all $\mb{x} \in \R^{n}$: 
    \begin{equation} \label{eq:issf_cbf_constraint}
        \sup_{\mb{v} \in \R^{m}}
       L_\mb{f}h(\mb{x},\bs{\rho}) + L_\mb{g}h(\mb{x},\bs{\rho})\mb{v}-\issfparam\Vert L_\mb{g}h(\mb{x},\bs{\rho}) \Vert^2 > -\alpha(h(\mb{x},\bs{\rho})).
    \end{equation}
\end{definition}

The parameter $\bs{\rho}\in\R^p$ contains information about the system's environment that affects safety, such as the location and size of obstacles. In novel environments the system may need to generate estimates of $\bs{\rho}$ denoted by $\widehat{\bs{\rho}}\in\R^p$ from complex measurements, such as camera data. The process of converting complex measurements to environmental parameters $\widehat{\bs{\rho}}$ is often imperfect, leading to error between the estimated and true values (i.e., $\widehat{\bs{\rho}} \neq \bs{\rho}$), which can cause safety violations. In this setting, safety can be achieved via {\em Measurement-Robust Control Barrier Functions (MR-CBFs)}:
\begin{definition}[Measurement-Robust Control Barrier Functions \citep{dean2020guaranteeing}]
The function $h$ is a \textit{Measurement-Robust Control Barrier Function} (MR-CBF) for \eqref{eq:reduced_model} on $\C$ with parameters $a,b\in\R_{\geq 0}$ if there exists $\alpha\in\K_{\infty}^{\rm e}$ such that for all $\widehat{\bs{\rho}}\in\R^p$ and $\mb{x}\in\R^n$:
\begin{equation} 
    \label{eq:mr_cbf_constraint}
        \sup_{\mb{v} \in \R^{m}}
        L_\mb{f}h(\mb{x},\widehat{\bs{\rho}}) + L_\mb{g}h(\mb{x},\widehat{\bs{\rho}})\mb{v}
        - a - b\Vert\mb{v}\Vert> -\alpha(h(\mb{x},\widehat{\bs{\rho}})).
\end{equation}
\end{definition}
The following theorem summarizes the safety results achieved with these various types of CBFs:
\begin{theorem} \label{thm:old_cbfs}
Consider the set $\C$ defined in~(\ref{eq:safe_set}).
\begin{enumerate}
 \setlength\itemsep{0em}
\item If $h$ is a CBF for \eqref{eq:reduced_model} on $\C$, $\mb{d}(t) = \mb{0}$ for $t\in\R_{\geq 0}$ and $\widehat{\bs{\rho}} = \bs{\rho}$, then there exists a controller $\mb{k}$ such that \eqref{eq:closed_loop_system} is safe with respect to $\C$.

\item If $h$ is an ISSf-CBF for \eqref{eq:reduced_model} on $\C$ with parameter $\varphi$ and $\widehat{\bs{\rho}} = \bs{\rho}$, then there exists a controller $\mb{k}$ such that \eqref{eq:closed_loop_system} is ISSf with respect to $\C$ with  $\gamma(\delta) = -\alpha^{-1}(-\delta^2/(4\varphi))$ where $\alpha^{-1}\in\K_{\infty}^{\rm e}$.

\item Assume $L_{\mb{f}}h$, $L_{\mb{g}}h$, and $\alpha\circ h$ are Lipschitz continuous on their domains, and assume that $\Vert\widehat{\bs{\rho}}-\bs{\rho}\Vert\leq\epsilon$ for some $\epsilon\in\R_{\geq 0}$. Then there exists $\underline{a},\underline{b}\in\R_{\geq 0}$ such that if $h$ is an MR-CBF for \eqref{eq:reduced_model} on $\C$ with parameters $a,b\in\R_{\geq 0}$ satisfying $a\geq\underline{a}$ and $b\geq\underline{b}$, and $\mb{d}(t) = \mb{0}$ for $t\in\R_{\geq 0}$, then there exists a controller $\mb{k}$ such that \eqref{eq:closed_loop_system} is safe with respect to $\C$.
\end{enumerate}
\end{theorem}

%% file: Sections/AT_SecIV_Draft.tex
In this section we propose a design paradigm that leverages \newpbl\, to select parameters for a CBF-based controller that achieves performance and safety for a multi-layered control system. 


\underline{\textit{Multi-Layered System Dynamics:}} Many real-life engineering systems have high-dimensional state spaces and complex dynamics. Hence  control systems are often designed as a set of interconnected subsystems, such as a low-dimensional subsystem that provides reference signals capturing safe behavior and a high-dimensional subsystem that tracks these reference signals.
In particular, consider the following cascaded nonlinear control-affine system resulting as a modification of \eqref{eq:reduced_model}:
\begin{equation}
    \dot{\mb{x}} = \mb{f}(\mb{x})+\mb{g}(\mb{x})\bs{\kappa}(\bs{\xi}), \qquad
 \dot{\bs{\xi}} = \mb{f}_{\bs{\xi}}(\mb{x},\bs{\xi}) + \mb{g}_{\bs{\xi}}(\mb{x},\bs{\xi})\mb{u},
 \label{eq:full_system}
\end{equation}
with additional states $\bs{\xi} \in \R^{n_{\bs{\xi}}}$, control input $\mb{u} \in \R^{m_{\bs{\xi }}}$ and functions
$\bs{\kappa}: \R^{n_{\bs{\xi}}} \to \R^m$, $\mb{f}_{\bs{\xi }}:\R^n \times \R^{n_{\bs{\xi }}} \to \R^{n_{\bs{\xi }}}$, and
$\mb{g}_{\bs{\xi }}:\R^n \times \R^{n_{\bs{\xi }}} \to \R^{n_{\bs{\xi }} \times m_{\bs{\xi }}}$ assumed to be locally Lipschitz continuous on their domains. We note that the input $\mb{v}$ from \eqref{eq:reduced_model} was replaced by $\bs{\kappa}(\bs{\xi})$. These dynamics may represent Euler-Lagrange systems such as robots, where $\mb{x}$ reflects base position, $\bs{\xi}$ captures base velocities and joint positions and velocities, and the input $\mb{u}$ reflects the torques applied to the joints. 

Given this cascaded system, we utilize the low-dimensional subsystem to ensure that $\mathcal{C}$ is ISSf by making two assumptions. First, we assume the safe set $\C$ can be described as in \eqref{eq:safe_set}, such that it only depends on the states $\mb{x}$ and parameters $\bs{\rho}$, and not the states $\bs{\xi}$. For example, in the context of a robotic system, this assumption is justified if safety is described as keeping the base position of the robot away from obstacles. Second, we assume there exists a controller $\bs{\pi}:\R^n\times\R^{n_{\bs{\xi}}}\times\R^m\to\R^{m_{\bs{\xi}}}$ and $\contbound\in\R_{\geq 0}$ such that for any continuous, bounded signal $\mb{s}:\R_{\geq 0}\to\R^m$, the closed-loop system:
\begin{equation} \label{eq:lowLevel_closed_loop}
    \dot{\bs{\xi}} = \mb{f}_{\bs{\xi}}(\mb{x},\bs{\xi}) + \mb{g}_{\bs{\xi}}(\mb{x},\bs{\xi})\bs{\pi}(\mb{x},\bs{\xi},\mb{s}(t)),
\end{equation}
satisfies the following implication:
\begin{equation}
    \Vert \bs{\kappa}(\bs{\xi}(0)) - \mb{s}(0) \Vert \leq \contbound \implies \Vert \bs{\kappa}(\bs{\xi}(t)) - \mb{s}(t) \Vert \leq \contbound,\quad t\in\R_{\geq 0}.
\end{equation}
This assumption reflects that a separate controller may be designed for the high-dimensional dynamics to track well-behaved reference signals synthesized via the low-dimensional model.
In particular, if a continuous controller $\mb{k}:\R^n\to\R^m$ is designed for the low-dimensional system \eqref{eq:reduced_model} and $\Vert\bs{\kappa}(\bs{\xi}(0))-\mb{k}(\mb{x}(0)) \Vert \leq \contbound$, then we have that the controller $\bs{\pi}$ ensures $\Vert\bs{\kappa}(\bs{\xi}(t))-\mb{k}(\mb{x}(t))\Vert \leq \contbound$ for $t\in\R_{\geq 0}$. With this assumption in mind, we may study the ISSf behavior of the closed-loop system:
\begin{equation}
    \dot{\mb{x}} = \mb{f}(\mb{x})+\mb{g}(\mb{x})(\mb{k}(\mb{x})+\mb{d}(t)), \qquad
 \dot{\bs{\xi}} = \mb{f}_{\bs{\xi}}(\mb{x},\bs{\xi}) + \mb{g}_{\bs{\xi}}(\mb{x},\bs{\xi})\bs{\pi}(\mb{x},\bs{\xi},\mb{k}(\mb{x})),
\label{eq:closed_loop_full}
\end{equation}
with the disturbance defined as $\mb{d}(t) = \bs{\kappa}(\bs{\xi}(t))-\mb{k}(\mb{x}(t))$ satisfying $\Vert\mb{d}\Vert_\infty \leq \contbound$. 

\underline{\textit{Combined Robust CBFs for PBL:}}
We now combine the robustness properties of MR-CBFs and ISSf-CBFs to account for measurement uncertainty and the disturbance, $\mb{d}$, allowing us to make robust safety guarantees for the full system \eqref{eq:closed_loop_full}. This is formalized in the following theorem:

\begin{theorem}\label{thm:all_uncertainties}
Given the set $\mathcal{C}$ defined in \eqref{eq:safe_set}, suppose the functions $L_\mb{f}h$, $L_\mb{g}h$, $\Vert L_\mb{g}h\Vert^2$, and $\alpha\circ h$ are Lipschitz continuous on their domains, and assume that $\Vert\widehat{\bs{\rho}}-\bs{\rho}\Vert \leq \epsilon$ for some $\epsilon \in\R_{\geq 0}$. Then there exists $\underline{a},\underline{b}\in\R_{\geq 0}$ such that if $h$ satisfies:
\begin{equation} \label{eq:tr_constraint}
    \sup_{\mb{v}\in\R^m}    L_\mb{f}h(\mb{x},\widehat{\bs{\rho}}) + L_\mb{g}h(\mb{x},\widehat{\bs{\rho}})\mb{v} - \issfparam\Vert L_\mb{g}h(\mb{x},\widehat{\bs{\rho}}) \Vert^2 - a - b\Vert\mb{v}\Vert > -\alpha(h(\mb{x},\widehat{\bs{\rho}})),
    \end{equation}
for all $\mb{x}\in\R^n$ and some $a,b\in\R_{\geq 0}$ satisfying $a\geq\underline{a}$ and $b\geq\underline{b}$, then there exists a controller $\mb{k}:\R^n\to\R^m$ such that \eqref{eq:closed_loop_full} is ISSf with respect to $\mathcal{C}$ with $\gamma(\delta) = -\alpha^{-1}(-\delta^2/(4\issfparam))$. 
\end{theorem}
\noindent 
The proof of this theorem 
can be found in the 
extended version of this paper \citep{extendedVersion}.
As in \cite{gurriet2018towards}, \eqref{eq:tr_constraint} can be incorporated as a constraint into a safety filter on a locally Lipschitz continuous nominal controller $\mathbf{k}_\textrm{nom}:\R^n \to \R^m$. We call this filter the Tunable Robustified Optimization Program \eqref{eq:new_controller} with tunable parameters $\alpha, \issfparam, a,$ and $ b$. 
\begin{align}\tag{TR-OP}
\label{eq:new_controller}
\mb{k}(\mb{x}) = {\rm arg}\!\!\min_{\mb{v} \in \R^m } & \quad \Vert \mb{v} - \mb{k}_{\rm nom}(\mathbf{x})  \Vert^2  \\
   \textrm{s.t. } & L_\mb{f}h(\mb{x}, \widehat{\bs{\rho}}_i) + L_\mb{g}h(\mb{x},  \widehat{\bs{\rho}}_i) \mb{v} - \issfparam\Vert L_\mb{g}h(\mb{x},\widehat{\bs{\rho}}_i) \Vert^2 - a - b\Vert\mb{v}\Vert \geq - \alpha h(\mb{x},  \widehat{\bs{\rho}}_i), \nonumber\\
   & \quad \quad \quad \quad \quad \quad \quad \quad \quad \quad \quad \quad \quad \quad \quad \quad \quad \quad \quad \quad \quad \quad \quad  \quad \forall i \in \{1, \dots, N_{o}\}.  \nonumber
\end{align}
Here we use a linear class~$\K_{\infty}^{\rm e}$ function with coefficient $\alpha \in \R_{>0}$. If we wish to enforce multiple safety constraints, such as in obstacle avoidance with several obstacles, 
 $\widehat{\bs{\rho}}_i$ can be used to indicate the measured parameters of the $i^{th}$ obstacle, with $N_{o} \in \mathbb{N}$ being the total number of obstacles. Enforcing this constraint for $N_o>1$ can be viewed as Boolean composition of safe sets \citep{glotfelter2018boolean}. 
Additionally, this safety filter is a Second-Order Cone Program (SOCP) \citep{boyd2004convex} for which an array of solvers exist including ECOS \citep{domahidi2013ecos}. 

\underline{\textit{Integrating Learning to Tune the Control Barrier Function:}}
The parameter selection process of \ref{eq:new_controller} is particularly important, since the parameters $\underline{a}$ and $\underline{b}$ guaranteed to exist by Theorem \ref{thm:all_uncertainties} are worst-case approximations of the uncertainty generated using Lipschitz constants. Such approximations often lead to undesired conservatism and may render the system incapable of performing its goal (as seen in Figure \ref{fig:trajectories}). 
Thus, as illustrated in Figure \ref{fig:overview}, we propose utilizing \newpbl\, to identify user-preferred parameters of \ref{eq:new_controller}. This relaxes the worst-case over-approximation to experimentally realize performant and safe behavior. This design paradigm relies on the tunable construction of \ref{eq:new_controller}, allowing us to define the actions for \newpbl\, to $\act = (\alpha, \issfparam, a, b)$. We note the construction of \ref{eq:new_controller} assures that unsafe actions are not necessarily catastrophic, as any $\alpha, \issfparam, a, b > 0$ endows the system with a non-zero degree of robustness to disturbances and measurement error. This assurance allows us to utilize a safety-aware approach where unsafe actions are considered undesirable as opposed to more conservative safety-critical approach to learning where unsafe actions are considered catastrophic.

%% file: Sections/experiments.tex

\begin{figure}[t] 
    \centering
    \includegraphics[width=0.75\linewidth]{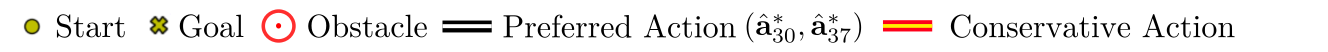}
    \includegraphics[width=0.32\linewidth]{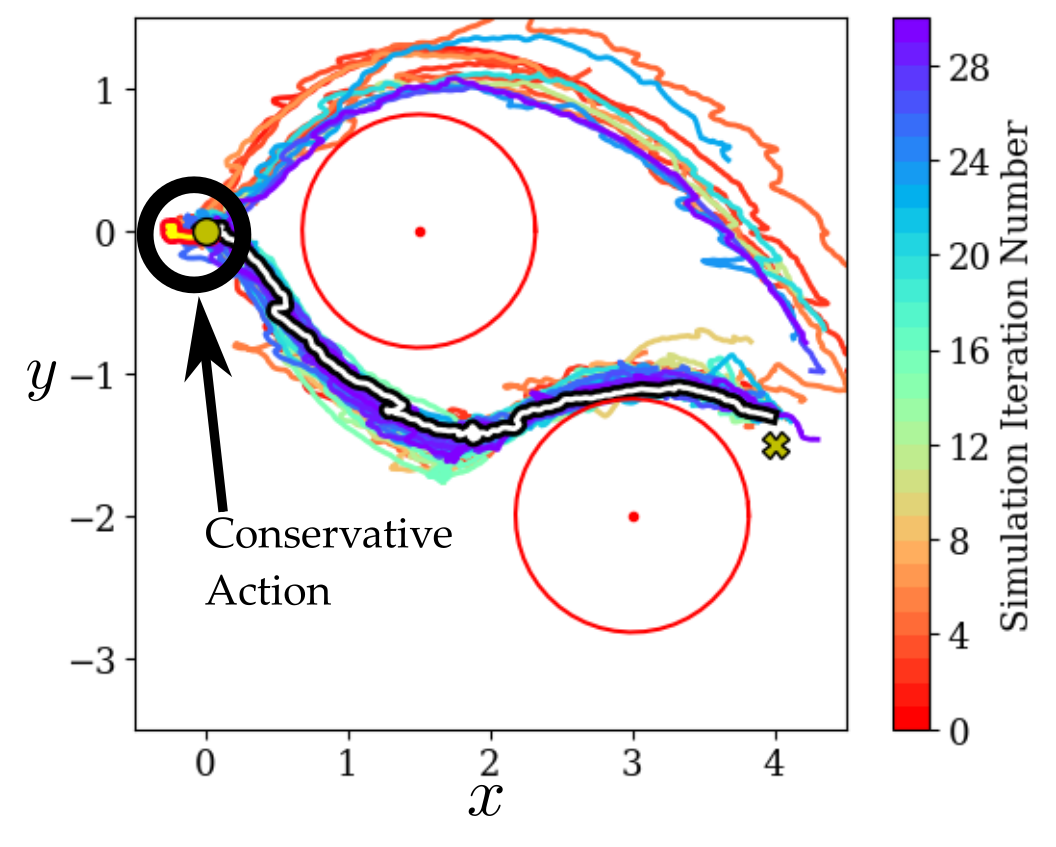}
    \includegraphics[width=0.32\linewidth, height=3.9cm]{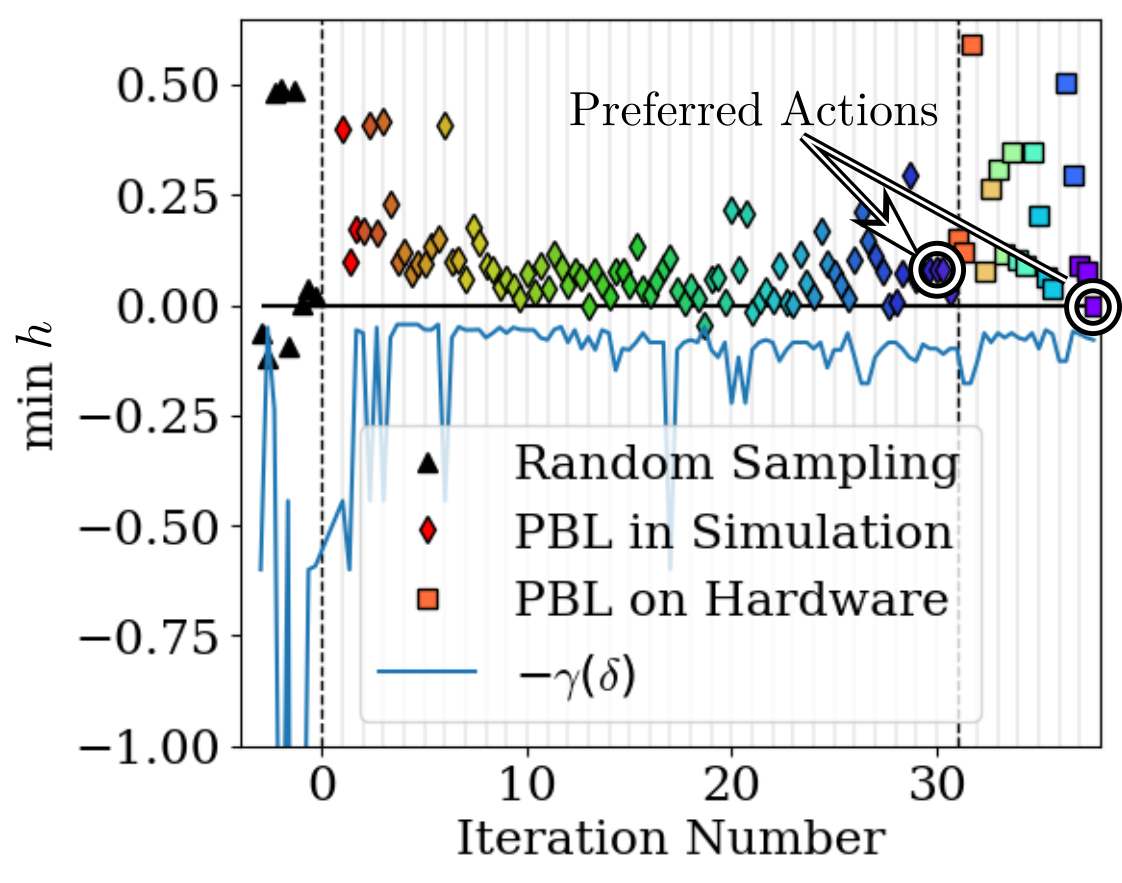}
    \includegraphics[width=0.32\linewidth]{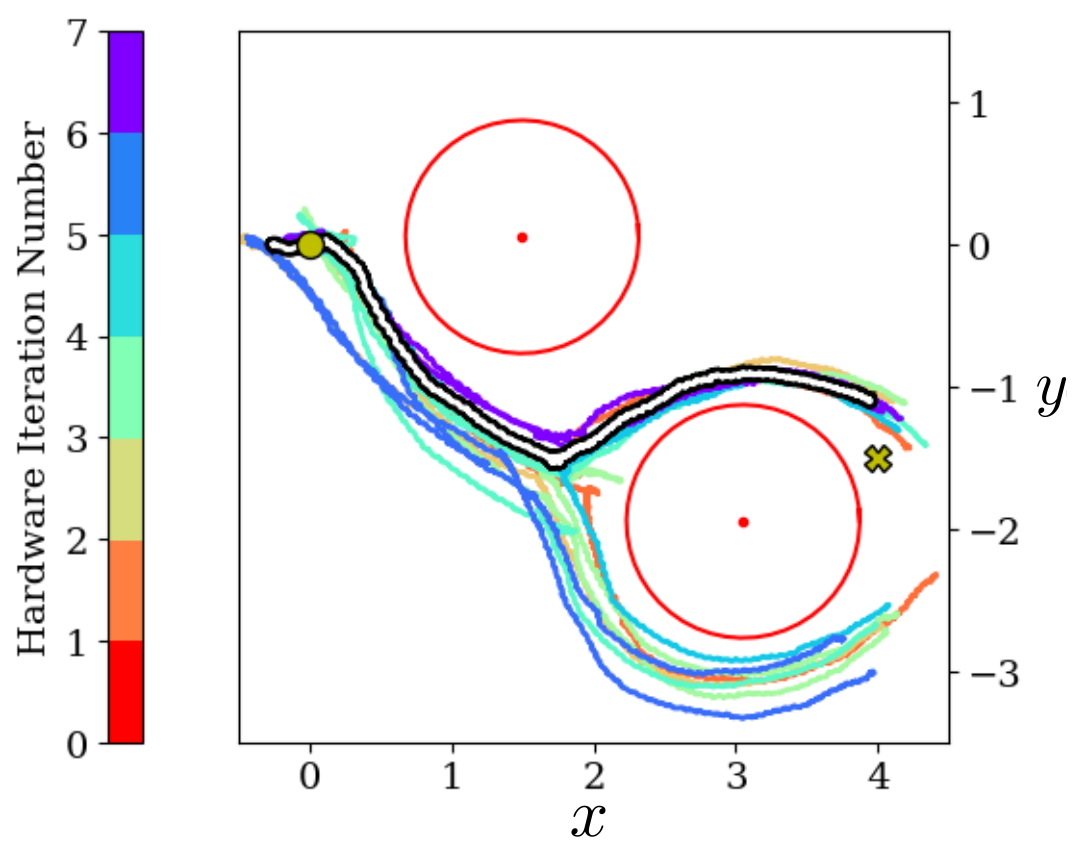}
    \caption{\textbf{(Left)} Actions sampled during simulation in 30 iterations with 3 new actions in each iteration.
    The preferred action,  $\hat{\mb{a}}_{30} = (3, 0.6, 0.5, 0.015)$, is shown in black and white. A conservative action, $\mb{a}= (2, 0.5, 0.0651, 0.485)$, is indicated by the black circle, where $a$ and $b$ were determined by estimating the Lipschitz coefficients present in the proof of Theorem~\ref{thm:all_uncertainties}. The conservative action fails to progress whereas \newpbl\ provides an action which successfully navigates between obstacles.
    \textbf{(Center)} The minimum value of $h$ that occurred in each iteration. Triangles, diamonds, and squares represent actions that are sampled randomly, by PBL in simulation and on hardware in an indoor setting, respectively. Colors correlate to iteration number. The lower bound $-\gamma(\delta) $ for the expanded set $\mathcal{C}_\delta$ with $\delta = 1$ is plotted. The preferred actions for simulation and hardware experiments are circled. \textbf{(Right)} Seven additional iterations of 3 actions executed indoors. The preferred action, $\hat{\mb{a}}_{37}^* = ( 4, 0.6, 0.4, 0)$, successfully traverses between the obstacles. 
    }
    \label{fig:trajectories}
    \vspace{-0.75cm}
\end{figure}

\begin{figure}[t]
\begin{minipage}[t]{0.7\linewidth}
\begin{figure}[H]
    \centering
    \includegraphics[width=\linewidth]{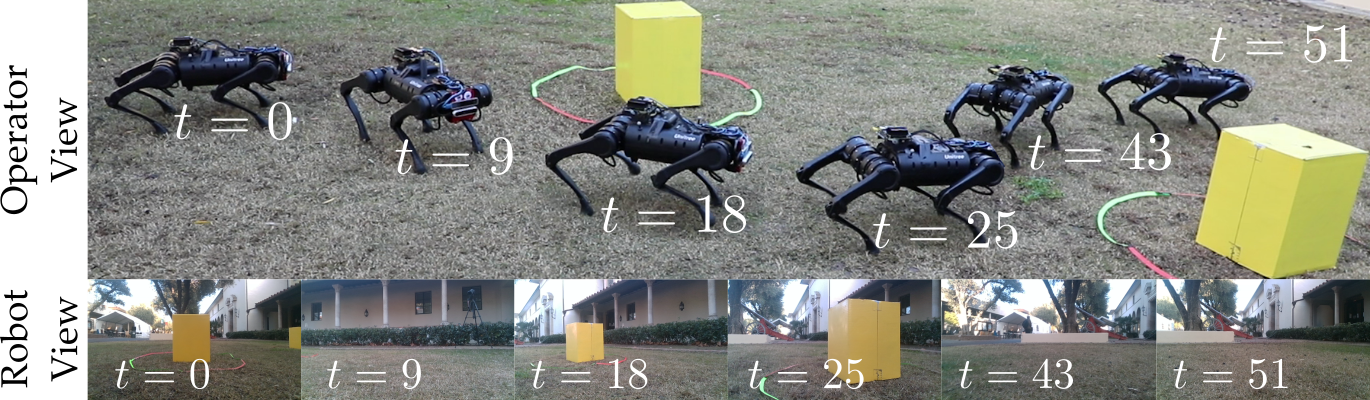}
    \caption{The preferred action,  $\hat{\mb{a}}_{40}^* = (5, 0.1, 0.4, 0.02)$, after simulation, indoor experiments, and 3 additional iterations of 3 actions in an outdoor environment is shown alongside views from the onboard camera.}
    \label{fig:outdoor}
\end{figure}
\end{minipage}
\;\;
\begin{minipage}[t]{0.27\linewidth}
\begin{table}[H]
    \centering
    \scriptsize
    \begin{tabular}{|c|c|}
        \hline  hyperparameter & value  \\
         \hline  $\lambda$ & $ -0.5$\\
         \hline $\beta $ & 0 \\
         \hline
    \end{tabular}
    \begin{tabular}{|c|c|c|c|}    
        \hline  name & min. & max. & $\Delta$  \\
         \hline $\alpha$ & 0.5 & 5 & 0.5 \\
         \hline $\issfparam$ & 0 & 1 & 0.1 \\
         \hline $a$ & 0 & 1 & 0.1 \\
         \hline $b$ & 0 & 0.05 & 0.005 \\
         \hline
    \end{tabular}
    \caption{The safety-aware hyperparameters, and action space bounds (min. and max.) with discretizations $\Delta$.}
    \label{tbl:pbl_params}
\end{table}
\end{minipage}
\vspace{-0.8cm}
\end{figure}

We applied the proposed design paradigm to a perception-based obstacle avoidance task with a Unitree A1 quadrupedal robot (Figure~\ref{fig:overview}) in  simulation and on hardware for both indoor and outdoor environments (see video: \cite{videoVideo}). The action space $\actspace$ and hyperparameters of PBL are defined in Table~\ref{tbl:pbl_params}. We used the unicycle model as our simplified model \eqref{eq:reduced_model} with the nominal controller $\mathbf{k}_\textrm{nom}$: 
\begin{align}
    \underbrace{\lmat \dot x \\ \dot y \\ \dot \psi \rmat}_{\dot{\mb{x}}}  =\underbrace{\lmat 0 \\ 0 \\ 0 \rmat}_{\mb{f}(\mb{x})} +  \underbrace{\lmat \cos\psi & 0 \\ \sin \psi & 0 \\ 0 & 1 \rmat}_{\mb{g}(\mb{x})}\left( \underbrace{\lmat v \\ \omega  \rmat}_{\mb{v}} + \mathbf{d}(t)\right), &&
    \mb{k}_\textrm{nom}(\mb{x}) =
    \lmat K_v d_\textrm{g} + C \\ - K_\omega (\sin\psi - (y_\textrm{g} - y)/d_g) \rmat,
    \label{eq:uni_and_nom_controller}
\end{align}
where $(x,y)$ is the planar position of the robot, $\psi$ is the yaw angle, $(x_\textrm{g}, y_\textrm{g}) $ is the goal position of the robot, $d_\textrm{g} = \Vert (x_\textrm{g} -x, y_\textrm{g} - y) \Vert$ is the distance to the goal, and $K_v, K_\omega$, and $C$ are positive constants. Obstacle avoidance is encoded via the 0-superlevel set of the function:
\begin{equation}
    h(\mb{x},\bs{\rho}_i) = d_{\textrm{obs},i} - r_{\textrm{obs}} - \zeta\cos(\psi - \theta_{i}) ,
\end{equation}
where $\bs{\rho}_i = [x_{\textrm{obs},i}, y_{\textrm{obs},i}]$ is the location of the $i^{th}$ obstacle, $d_{\textrm{obs},i} = \Vert (x_{\textrm{obs},i} - x, y_{\textrm{obs},i} - y)\Vert$ and 
$\theta_{i} = \arctan((y_{\textrm{obs},i}-y)/(x_{\textrm{obs},i} - x))$ are the distance and angle from the $i^{th}$ obstacle, 
$r_{\textrm{obs}}$ is the sum of the radii of the obstacle and robot, and $\zeta>0$ determines the effect of the heading angle on safety. The controller used to drive the system is the \ref{eq:new_controller} with the nominal controller $\mathbf{k}_\textup{nom}$ from \eqref{eq:uni_and_nom_controller}. In practice, infeasibilities of this safety filter were considered unsafe and the inputs were saturated such that $v \in [-0.2, 0.3]\,\textrm{m/s}$ and $\omega \in [-0.4, 0.4]\,\textrm{rad/s}$. The velocity command $\mb{v}$ is computed at 20 Hz and error introduced by this sampling scheme is captured by the tracking error $\mb{d}(t)$. Tracking of $\mb{v}$ is performed by an inverse dynamics quadratic program (ID-QP) walking controller designed using the concepts in \cite{buchli2009compliant}, which realizes a stable walking gait for \eqref{eq:closed_loop_full} at 1 kHz.

\underline{\textit{Simulation results:}} We simulated the quadruped executing the proposed controller with parameters provided by \newpbl. The resulting trajectories and the position of the obstacles are shown in Figure \ref{fig:trajectories}. 
We ran 30 iterations, with 3 new actions sampled in each iteration ($s = 3$), and obtained user preferences and ordinal labels in between each set of actions. To simulate perception error, the measurements of the obstacles were shifted by $-0.1 $ m in the $y$-direction.
The parameters found with \newpbl\ allow the robot to navigate between obstacles.
For comparison, a conservative action 
is also shown, which is safe but fails to progress towards the goal.
\newpbl \, eliminates this conservatism with only minor safety violations and determines a parameter set which is both safe and performant.

\underline{\textit{Hardware results:}} After simulation, we continued learning on hardware experiments in a laboratory setting for 7 additional iterations until the user was satisfied with the experimental behavior. 
The robot and obstacle positions were estimated using Intel RealSense T265 and D415 cameras to perform SLAM and segmentation. Centroids of segmented clusters in the occupancy map were used as the measured obstacle positions $\widehat{\bs{\rho}}_i$.
The true robot and obstacle positions were obtained for comparison using an OptiTrack motion capture system. The results of these experiments can be seen in Figure \ref{fig:trajectories}.
Afterwards, three additional iterations were conducted outdoors on grass until again the user was satisfied with the experimental behavior. The resulting best trajectory can be seen in Figure \ref{fig:outdoor}. The preferred action was also tested on a variety of other obstacle arrangements to confirm its generalizability. The performance of the final preferred action for these obstacle configurations can be seen in the supplementary video \citep{videoVideo}.




\section{Conclusion}
In this work we proposed a design paradigm for control systems in which the robust safety requirements of a provably safe, but conservative controller are relaxed, and controller parameters are instead chosen using a Preference-Based Learning algorithm called \newpbl. Using our algorithm, we were able to learn a set of parameters that leads to user-preferred balance between safety and robustness on a quadrupedal robot platform. Future work includes applying this framework to other platforms such as bipedal robots, autonomous vehicles, and assistive devices, and to more complicated environments like obstacles with time-varying parameters.

%% file: Sections/Appendices/theoremProof.tex
In this appendix we present the proof of Theorem \ref{thm:all_uncertainties} which establishes the input-to-state safety of $\mathcal{C}$ in the context of disturbances and measurement uncertainties. Recall that Theorem \ref{thm:all_uncertainties} is given as: 


\begin{theorem*}\textbf{\textup{\ref{thm:all_uncertainties}}}
Given the set $\mathcal{C}$ defined in \eqref{eq:safe_set}, suppose the functions $L_\mb{f}h$, $L_\mb{g}h$, $\Vert L_\mb{g}h\Vert^2$, and $\alpha\circ h$ are Lipschitz continuous on their domains, and assume that $\Vert\widehat{\bs{\rho}}-\bs{\rho}\Vert \leq \epsilon$ for some $\epsilon \in\R_{\geq 0}$. Then there exists $\underline{a},\underline{b}\in\R_{\geq 0}$ such that if $h$ satisfies:
\begin{equation} \label{eq:tr_constraint_appdx}
    \sup_{\mb{v}\in\R^m}    L_\mb{f}h(\mb{x},\widehat{\bs{\rho}}) + L_\mb{g}h(\mb{x},\widehat{\bs{\rho}})\mb{v} - \issfparam\Vert L_\mb{g}h(\mb{x},\widehat{\bs{\rho}}) \Vert^2 - a - b\Vert\mb{v}\Vert > -\alpha(h(\mb{x},\widehat{\bs{\rho}})),
    \end{equation}
for all $\mb{x}\in\R^n$ and some $a,b\in\R_{\geq 0}$ satisfying $a\geq\underline{a}$ and $b\geq\underline{b}$, then there exists a controller $\mb{k}:\R^n\to\R^m$ such that \eqref{eq:closed_loop_full} is ISSf with respect to $\mathcal{C}$ with $\gamma(\delta) = -\alpha^{-1}(-\delta^2/(4\issfparam))$ where $\alpha^{-1}\in\K_{\infty}^{\rm e}$. 
\end{theorem*}

\noindent
\begin{proof}
First, we show that satisfying \eqref{eq:tr_constraint_appdx} for a particular set of $\underline{a}, \underline{b}$ implies satisfaction of \eqref{eq:issf_cbf_constraint}.
For this we choose: 
\begin{align}
    \underline{a}  = \epsilon(\mathcal{L}_{L_\mb{f}h} + \mathcal{L}_{\alpha \circ h } + \mathcal{L}_{\issfparam \Vert L_\mb{g}h \Vert^2} ), && 
    \underline{b}  = \epsilon \mathcal{L}_{L_\mb{g}h},
\end{align}
where $\mathcal{L}$ indicates the Lipschitz coefficient of the subscripted function with respect to argument $\bs{\rho}$.
Let us define the function $c:\R^n \times \R^p \times \R^m \to \R$ such that:
\begin{equation}
    c(\mb{x}, \bs{\rho}, \mb{v}) \triangleq L_\mb{f}h(\mb{x},\bs{\rho}) + L_\mb{g}h(\mb{x},\bs{\rho}) \mb{v} - \issfparam \Vert L_\mb{g}h(\mb{x},\bs{\rho})\Vert^2 + \alpha(h(\mb{x},\bs{\rho})).
\end{equation}
Using this definition we have for all $\mb{v} \in \R^m$ that:
\begin{align}
    c(\mb{x}, \bs{\rho}, \mb{v})
    & = c(\mb{x}, \widehat{\bs{\rho}}, \mb{v}) + c(\mb{x}, \bs{\rho}, \mb{v})  - c(\mb{x}, \widehat{\bs{\rho}}, \mb{v}),
    \label{pf:add_zero}\\
    & \geq c(\mb{x}, \widehat{\bs{\rho}}, \mb{v}) - \underbrace{\epsilon (\mathcal{L}_{L_\mb{f}h} + \mathcal{L}_{\alpha \circ h } + \mathcal{L}_{\issfparam \Vert L_\mb{g}h \Vert^2})}_{\underline{a}} - \underbrace{\epsilon  \mathcal{L}_{L_\mb{g}h}}_{\underline{b}} \Vert \mb{v} \Vert  \label{pf:lipschitz}\\
    & \geq c(\mb{x}, \widehat{\bs{\rho}}, \mb{v}) - a - b\Vert \mb{v} \Vert.
    \label{pf:ab_bound}
\end{align}
Above we added zero in \eqref{pf:add_zero} and used the Lipschitz coefficients and the worst case uncertainty $\epsilon$ to achieve the bound in \eqref{pf:lipschitz}. 
Since $\sup_{\mb{v}\in\R^m} c(\mb{x}, \widehat{\bs{\rho}}, \mb{v}) - a - b\Vert\mb{v}\Vert > 0$ holds based on \eqref{eq:tr_constraint_appdx}, inequality \eqref{pf:ab_bound}
implies that \eqref{eq:issf_cbf_constraint} holds for the true parameters, $\bs{\rho}$. Since \eqref{eq:issf_cbf_constraint} holds, the conditions of Theorem~\ref{thm:old_cbfs} point 2 are satisfied and thus $\mathcal{C}$ is ISSf with $\gamma(\delta) = -\alpha^{-1}(-\delta^2/(4\issfparam))$ where $\alpha^{-1} \in \K_{\infty}^{\rm e}$.
\end{proof}